\documentclass[conference]{IEEEtran}

\IEEEoverridecommandlockouts
\usepackage{cite}
\usepackage{amsmath,amssymb,amsfonts}
\usepackage{algorithmic}
\usepackage{graphicx}
\usepackage{booktabs}
\usepackage{textcomp}
\usepackage{xcolor}
\usepackage[nolist]{acronym}
\usepackage{multirow}
\def\BibTeX{{\rm B\kern-.05em{\sc i\kern-.025em b}\kern-.08em
    T\kern-.1667em\lower.7ex\hbox{E}\kern-.125emX}}
    
\usepackage{lipsum}
\newcommand{\nota}[1]{\textcolor{blue}{#1}}
\let\olip\lipsum
\renewcommand{\lipsum}[1][]{\nota{\olip[#1]}}

\usepackage{epstopdf}
\begin{document}

\begin{acronym}
    \acro{AP}{Arbitrary Precision}
    \acro{DT}{Decision Trees}
    \acro{HLS}{High Level Synthesis}
    \acro{ML}{Machine Learning}
\end{acronym}

\title{A Flexible HLS Hoeffding Tree Implementation for Runtime Learning on FPGA}

\author{
    \IEEEauthorblockN{Luís Miguel Sousa}
    \IEEEauthorblockA{\textit{Faculty of Engineering} \\
        \textit{University of Porto}\\
        Porto, Portugal \\
        lm.sousa@fe.up.pt}
    \and
    \IEEEauthorblockN{Nuno Paulino}
    \IEEEauthorblockA{\textit{INESC-TEC and Faculty of Engineering} \\
        \textit{University of Porto}\\
        Porto, Portugal \\
        nuno.m.paulino@inesctec.pt}
    \and
    \IEEEauthorblockN{João Canas Ferreira, João Bispo}
    \IEEEauthorblockA{\textit{INESC-TEC and Faculty of Engineering} \\
        \textit{University of Porto}\\
        Porto, Portugal \\
        \{jcf,~jbispo\}@fe.up.pt}
}

\maketitle

\begin{abstract}
    Decision trees are often preferred when implementing Machine Learning in embedded systems for their simplicity and scalability. Hoeffding Trees are a type of Decision Trees that take advantage of the Hoeffding Bound to allow them to learn patterns in data without having to continuously store the data samples for future reprocessing. This makes them especially suitable for deployment on embedded devices.
    In this work we highlight the features of an HLS implementation of the Hoeffding Tree. The implementation parameters include the feature size of the samples (D), the number of output classes (K), and the maximum number of nodes to which the tree is allowed to grow (Nd).
    We target a Xilinx MPSoC ZCU102, and 
    evaluate: the design's resource requirements and clock frequency for different numbers of classes and feature size, the execution time on several synthetic datasets of varying sample sizes (N), number of output classes and the execution time and accuracy for two datasets from UCI.
    For a problem size of D3, K5, and N40000, 
    a single decision tree operating at 103MHz is capable of 8.3\texttimes\ faster inference than the 1.2\,GHz ARM Cortex-A53 core.
    Compared to a reference implementation of the Hoeffding tree, we achieve comparable classification accuracy for the UCI datasets.
\end{abstract}


\section{Introduction}

With the rise of edge computing, FPGA vendors have been releasing and marketing CPU\texttt{+}FPGA SOCs as the ideal solution for this domain. As edge devices are often specialised for a single task in a constrained environment, it is advantageous to build dedicated hardware to improve performance and energy efficiency. FPGAs offer the advantage of targeted hardware without losing the ability to adapt the platform to changes (e.g., security updates), while being more efficient than a pure software solution.

As \ac{HLS} matures~\cite{XilinxInc.2020VivadoSynthesis}, it becomes a more attractive approach to creating efficient high-preformance accelerators for FPGA devices.

\ac{ML} algorithms are a prime candidate for acceleration at the edge, but their computational requirements exceed the capabilities of many embedded devices. Inference at the edge is a problem being addressed by many works, but training at the edge still faces hurdles to adoption despite its clear benefits. In the field of \acp{DT}, many algorithms are incompatible with devices of this class due to memory constraints.
ID3~\cite{Quinlan1983LearningGames}, and derivatives such as C4.5 and C5.0 require the entire training dataset be present in memory for training. Incremental learning algorithms as ID5~\cite{UTGOFF1988ID5:ID3}, ID5R~\cite{Utgoff1989IMPROVEDLEARNING} and ITI~\cite{Utgoff1997DecisionRestructuring} do allow for ongoing learning from streaming data but store the dataset samples within the tree.

Hoeffding Trees ~\cite{Domingos2000MiningStreams} are incremental learning trees, which are more suitable for embedded scenarios because they have the following advantages: They asymptotically guarantee the same classification as traditional batch learners, and they store information about the distribution of samples statistically rather than the samples themselves, which drastically reduces memory requirements, especially for large datasets.

In this work, we present a flexible C/C\texttt{++} \ac{HLS} implementation of a Hoeffding Tree variant tailored for use in FPGAs, originally proposed by Lin et al.~\cite{Lin2019TowardsFPGA}.
Their work built on an earlier variant in which the storage of the statistical data of the sampling distribution of the original Hoeffding Tree was replaced by a Gaussian approximation~\cite{Pfahringer2008HandlingTrees}. Lin et al. replace this approximation with quantile estimation using asymmetric signum functions~\cite{Althoff2017AnTesting}. The result is a larger memory footprint but a reduction in computational requirements, while achieving similar results. Since it is implemented in Verilog, the applicability of the implementation is limited to circuit synthesis, e.g. for FPGA. By using \ac{HLS}, an implementation can be created that is equally suitable for CPU and FPGA.

The contributions of this work are as follows:
\begin{itemize}
    \item A generic, template-based C/C++ implementation of the Hoeffding Tree classifier as per Lin et al. \cite{Lin2019TowardsFPGA}, but that is suited for \ac{HLS}.
    \item Functional validation of the implementation through software execution, and post-synthesis onto a Xilinx ZCU102 development board.
    \item Experimental evaluation of memory requirements of the tree object as a function of template parameters.
    \item Experimental evaluation of FPGA resource requirements and execution time of the synthesised training and inference method as a function of template parameters.
\end{itemize}

\section{HLS Hoeffding Tree Implementation}

A decision tree is a type of machine learning algorithm used either for classification or regression.
A decision tree performs sequential binary decisions over an incoming vector of features, and a classification is computed when a leaf node is reached. During training, leaf nodes are added to the tree based on a splitting criteria, which separates the data into two regions at every tree junction. 
A Hoeffding tree is a type of decision tree where the criteria is the Hoeffding bound, shown in Equation \ref{eq1}. 
The tree performs learning and inference by relying on a property of the Hoeffding bound that guarantees that best splitting point is chosen. If a gain function $G$, is to be maximised, then given $G(X)$ and $G(Y)$ (X and Y being the attributes that generate the highest and second highest values of $G$) if $G(X)-G(Y)>\varepsilon$ then the Hoeffding bound guarantees that with probability $1-\delta$ X is the best attribute to split on. $R$ represents the range of the attributes e $N$ the number of samples on a node.

\begin{equation}
    \varepsilon = \sqrt{\frac{R^2ln(1/\delta)}{2N}}
    \label{eq1}
\end{equation}

\smallskip

Over other criteria, the Hoeffding bound has two characteristics: it allows for online incremental learning and growth of the tree which asymptotically tends towards the results provided by batch learners, and is independent of the probability distribution of the data sampling. The Hoeffding tree allows for continuous learning and node splitting for a potentially infinite (e.g. streaming applications) number of samples \cite{Domingos2000MiningStreams}.

FPGAs have been intensively studied for decision tree implementations, as a tree structure maps efficiently to specialised hardware. In conjunction with other optimisations, decision trees in FPGAs have been shown to outperform CPU and GPU solutions \cite{Barbareschi2021AdvancingStudy}.
Lin et al. \cite{Lin2019TowardsFPGA} demonstrate speedups of up to 1500x for an RTL implementation of the Hoeffding tree versus a 2.6GHz processor. Our aim is to explore a higher abstraction level via HLS, providing greater applicability features, while evaluating the attainable performance.

\begin{figure}[t]
    \centerline{\includegraphics[width=0.9\linewidth]{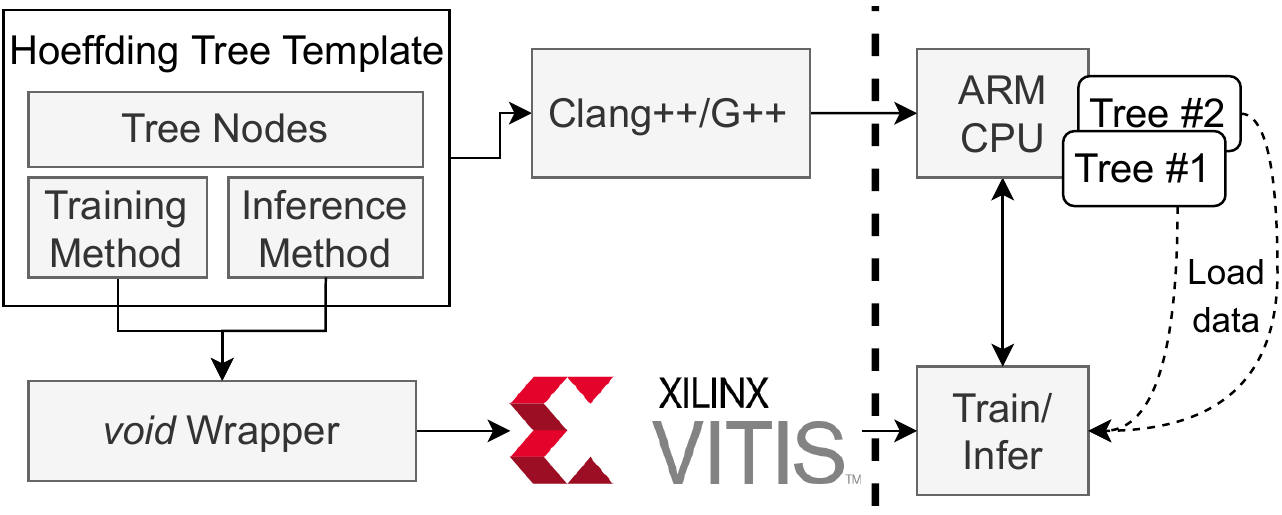}}
    \caption{Software and hardware architecture of the Hoeffding Tree implementation; the training and inference kernels are shared by multiple tree objects}
    \label{fig:software}
\end{figure}

We implemented the tree as a C\texttt{++} class template. The parameters include the maximum number of nodes in the tree, the feature size, and the floating-point precision. The class contains the training and inference methods which are synthesised to hardware. At runtime, the C\texttt{++} tree object can be manipulated in software, and passed as an argument to the training/inference method, as summarised in Figure \ref{fig:software}.

This allows for instantiation of several tree objects in memory (with different template parameters if desired). Trees with the same template parameters can be processed by the same synthesised circuit. Since the functions can also be invoked in software, this means that training or inference can be dynamically partitioned based on which device performs better for either task, as a function of the tree parameters. This also means that if FPGA is occupied processing a tree object, other trees can be evaluated via software without the need for a blocking wait. 

Finally, evaluation of multiple trees is possible by either a combination of software and hardware invocations, by deploying multiple instances of the hardware kernel, or by time-multiplexing a single hardware kernel (as explained below). Either case allows for the possibility of arbitrary runtime tree ensembles. This evaluation is currently future work. 

The Xilinx Vitis HLS flow enforces an OpenCL model for kernel invocation. The implemented kernel, \texttt{krnl\_Tree}, receives 4 arguments. A \texttt{HoeffdingTree} object as mentioned, an array of samples, an array of output classifications, and the size of these arrays.

In this model, a large overhead penalty would occur for invocations with a single sample, due to the data transfer time. A practical application of the kernel design could be, e.g., in the sensor domain, where the tree could continuously sample fused data from multiple sensors (i.e., multiple attributes) without processor intervention, avoiding transfer overheads. Alternatively, streaming samples can be accumulated until a sufficiently large number is held that mitigates this overhead. This does not mean that the tree behaves as a batch learner, as one sample is processed per each \emph{infer-then-train} step.

Inference on an incremental learning decision tree cannot be easily parallelised as the model changes and evolves with every training sample that arrives. This restricts the pipeline to dealing with one sample at a time, sequentially.
The sample structure contains information about whether it should be used for training purposes or only for inference. Thus, as the kernel loops through the sample array, it executes either the \texttt{train} or \texttt{infer} method of the tree object accordingly. The results are placed in the output data structure.

The OpenCL API allows for fine-grained control of how these arguments are passed to the kernels, each argument being a separate buffer with persistent storage.
Thus, trees can be transferred to FPGA memory once, and not retrieved between executions of the kernels. 
With this mechanism, a tree object can reside in memory while only new samples are transferred in, and the model can be retrieved in a final stage. 

Conversely, the samples themselves may remain in memory, and trees freely exchanged. 
This is one strategy for the construction of tree ensembles mentioned previously. Trees can reuse the same kernel instance via time-multiplexing, or by concurrent instantiation of several copies of \texttt{krnl\_Tree}. In either case, the same read-only sample buffer can be assigned to all trees, thus significantly reducing overhead and preventing data duplication. For brevity, the evaluation of ensembles is out of the scope of this paper.

\section{Experimental Evaluation}

We performed the following experiments: evaluated the resource utilisation of a single synthesised tree for a range of values for the feature size and number of classes; evaluated the training and inference time of a single tree in hardware, versus the ARM CPU, for several synthetic clustering datasets (varying number of point, clusters, and feature size); evaluated the classification accuracy and execution time of a single tree for UCI's Bank and Covertype datasets.

\subsection{Resource Utilisation}

Table \ref{tab:resources} presents various configurations of the kernel, tailored for datasets of different dimensions (D), with different number of classes (K), number of samples (N) and max number of nodes (Nd). The purpose is to determine the effect of these parameters on FPGA resource utilisation.
As expected, parameter N has no effect on resource utilisation as samples cannot be processed in parallel.

The feature size and the number of classes result in an increase in resource usage. This is due to the highly sequential nature of the generated kernel, which also explains why the performance of this kernel on training tasks is poor compared to the CPU. This overall advantage is less surprising when considered in the context of an 11-fold CPU advantage in clock speed. Current \ac{HLS} tools cannot automatically parallelize sequential code. Without hardware design expertise in order to optimise the design, the implementation will be far from optimal. In our implementation, we still believe that further parallelization can be achieved even within a single tree, through inner loop unrolling or memory partitioning. 

One interesting result is that of the kernel's operating frequency. It remains unchanged for all configurations. Looking deeper into the cause of this phenomenon, one finds that the bottleneck is the sorting of a sample down from the root node to the appropriate leaf node. This sequential operation also prevents the kernel from being pipelined.

\begin{table*}[htb]
    \centering
    \caption{N, D, K and Nd effects on FPGA Resource Utilisation}
    \begin{tabular}{crrrrrrrrr}
    	\toprule
        Nodes & 100 & 100 & 100 & 1000 &  100 &  100 &  100 & 1000 \\
        K     &   5 &   5 &  10 &    5 &    5 &    5 &   10 &    5 \\
        D     &   3 & 100 &   3 &    3 &    3 &  100 &    3 &    3 \\
        N     & 40k & 40k & 40k &  40k & 500k & 500k & 500k & 500k \\
        \midrule
        LUT    & 23304 (8.6\%) & 20567 (7.6\%) & 23776 (8.8\%) & 24351 (9.0\%) & 23304 (8.6\%) & 20567 (7.6\%) & 23776 (8.8\%) & 24351 (9.0\%) \\
        
        \midrule
        LUTRAM & 1395 (1.0\%) & 1179 (0.8\%) & 1399 (1.0\%) & 1397 (1.0\%) & 1395 (1.0\%) & 1179 (0.8\%) & 1399 (1.0\%) & 1397 (1.0\%) \\

        \midrule
        FF     & 35682 (6.6\%) & 29775 (5.5\%) & 36374 (6.7\%) & 36336 (6.7\%) & 35682 (6.6\%) & 29775 (5.5\%) & 36374 (6.7\%) & 36336 (6.7\%) \\
        
        \midrule
        BRAM   & 12 (1.3\%) & 9.5 (1.0\%) & 12 (1.3\%) & 12 (1.3\%) & 12 (1.3\%) & 9.5 (1.0\%) & 12 (1.3\%) & 12 (1.3\%) \\
        \midrule
        DSP    & 23 (0.9\%) & 25 (1.0\%) & 25 (1.0\%) & 25 (1.0\%) & 23 (0.9\%) & 25 (1.0\%) & 25 (1.0\%) & 25 (1.0\%) \\
        \midrule
        BUFG   & 13 (3.2\%) & 13 (3.2\%) & 13 (3.2\%) & 13 (3.2\%) & 13 (3.2\%) & 13 (3.2\%) & 13 (3.2\%) & 13 (3.2\%) \\
        \midrule
        MMCM   & 1 (25.0\%) & 1 (25.0\%)  & 1 (25.0\%)  & 1 (25.0\%)  & 1 (25.0\%)  & 1 (25.0\%)  & 1 (25.0\%)  & 1 (25.0\%) \\
        \midrule
        Freq. (MHz)             & 103.6 & 103.6 & 103.6 & 103.6 & 103.6 & 103.6 & 103.6 & 103.6 \\
        \bottomrule
    \end{tabular}
    \label{tab:resources}
\end{table*}

\begin{figure}[h]
    \centerline{\includegraphics[width=\linewidth]{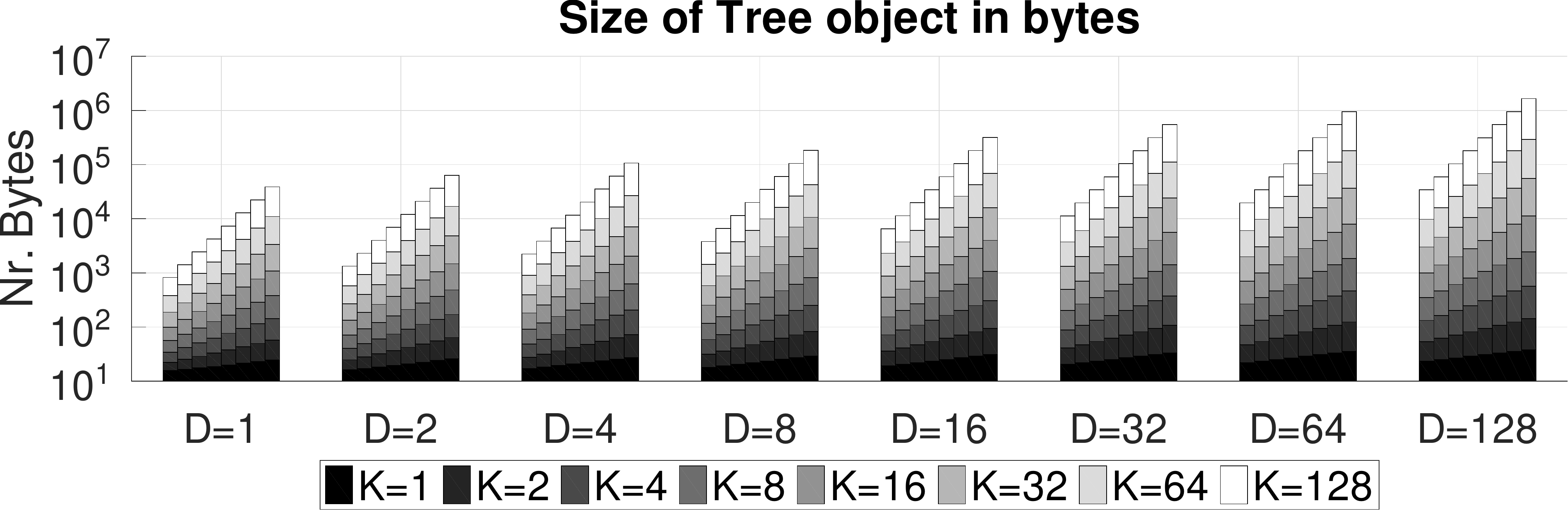}}
    \caption{Size of Tree objects in bytes for Nd, D and K. Each bar in every grouping, depicts a tree with a max number of nodes from $2^0$ to $2^7$.}
    \label{fig:bytes}
\end{figure}

\subsection{Performance}

These results were obtained by feeding the tree with datasets of K clusters in a D dimensional spaces, constituted of N points.
For these experimental runs, we will have the entire dataset transferred in a single operation to the FPGA's memory.

\begin{figure}[h]
    \centerline{\includegraphics[width=\linewidth]{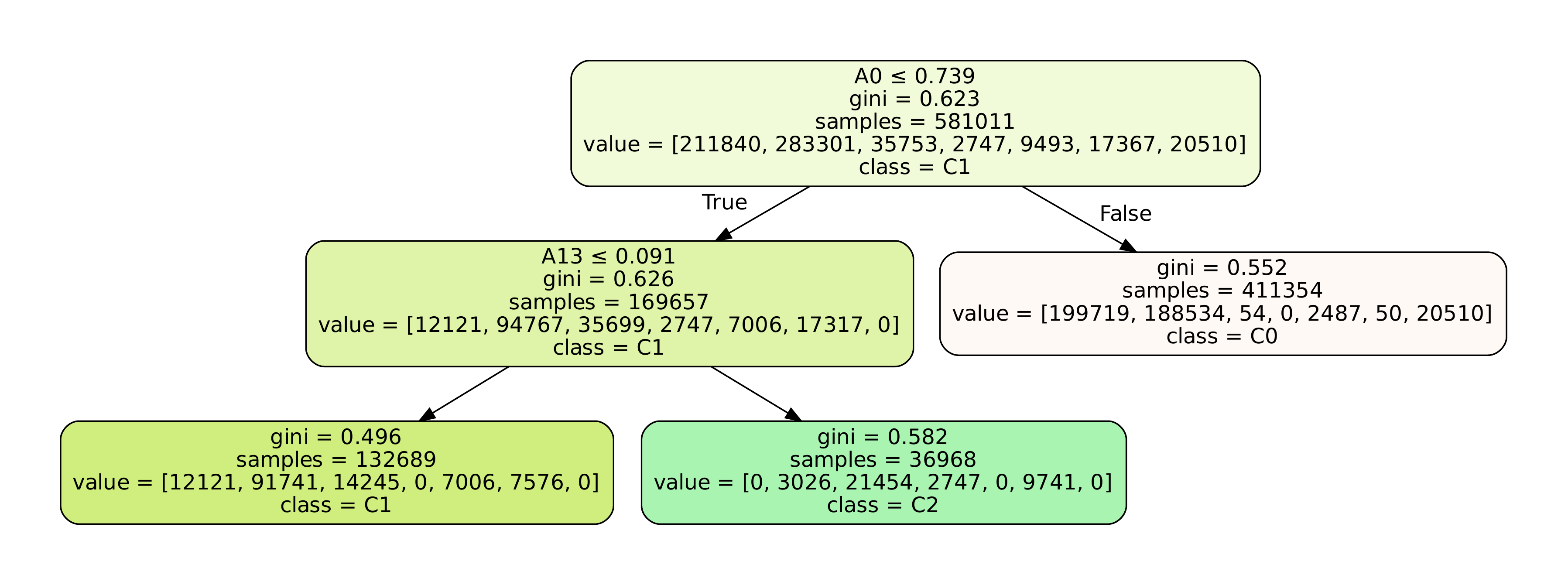}}
    \caption{Illustrative visualisation of tree model derived from UCI Covertype dataset. The tree was only allowed to grow to 5 nodes.}
    \label{fig:viz}
\end{figure}

Looking at the first four rows of Table \ref{tab:benchmarks} (D=3) it can be observed that for a 3-dimensional dataset, regardless of the bundle size, the ARM CPU in the ZCU102 SoC significantly outperforms the FPGA implementation in both the training and inference tasks. Also, the performance gap between both implementations grows with the number of samples processed. This indicates that the kernel is slower, per iteration, than the pure software solution.
Regarding the last four rows of Table \ref{tab:benchmarks} (D=100), the ARM CPU still outperforms the FPGA kernel in training. However, it does it with a lower margin and one that does not appear to grow with the added number of samples. On the inference task with this larger dataset, the FPGA outperforms the ARM processor by 8.3\texttimes.

Table \ref{tab:uci_datasets} presents benchmarks of two of the UCI datasets used by Lin et al. \cite{Lin2019TowardsFPGA}. The same tree parameters were used ($\delta=0.001$, $\lambda=0.01$, $\tau=0.05$, $n_{min}=200$, $n_{pt}=10$, $n_{quantiles}=16$, $Nd=2047$), with one being of special relevance: Nd (maximum number of nodes). A significant slowdown occurred. With the increased number of nodes, the sequential tree traversal algorithm increases in length. Our HLS implementation achieves comparable accuracy for \emph{Bank}, although the performance for \emph{Covertype} is inferior. Lin et al. \cite{Lin2019TowardsFPGA} reports 89.30\% and 72.51\%, respectively. We believe a difference in calculation precision between the CPU and FPGA caused the degradation, despite the use of 32-bit floating point data types for both devices.

\begin{table}[htb]
    \centering
    \caption{Training and inference times for four synthetic clustering datasets, for the ARM CPU (1.2Ghz) and the FPGA (103MHz)}
    \begin{tabular}{ccccrrr}
    \toprule
    K                          & D                          & N                             & Task   & ARM CPU    & FPGA       & Speedup    \\ \midrule
    \multirow{8}{*}[-2.2em]{5} & \multirow{4}{*}[-1em]{3}   & \multirow{2}{*}[-0.4em]{40k}  & Training   & 207 ms     & 1,990 ms   & 0.10\texttimes    \\ \cmidrule{4-7} 
                               &                            &                               & Inference  & 151 ms     & 462 ms     & 0.33\texttimes    \\ \cmidrule{3-7} 
                               &                            & \multirow{2}{*}[-0.4em]{500k} & Training   & 2,983 ms   & 30,933 ms  & 0.10\texttimes    \\ \cmidrule{4-7} 
                               &                            &                               & Inference  & 2,260 ms   & 11,442 ms  & 0.20\texttimes    \\ \cmidrule{2-7} 
                               & \multirow{4}{*}[-1em]{100} & \multirow{2}{*}[-0.4em]{40k}  & Training   & 6,028 ms   & 51,648 ms  & 0.12\texttimes    \\ \cmidrule{4-7} 
                               &                            &                               & Inference  & 3,924 ms   & 469 ms     & 8.37\texttimes    \\ \cmidrule{3-7} 
                               &                            & \multirow{2}{*}[-0.4em]{500k} & Training   & 75,763 ms  & 651,775 ms & 0.12\texttimes    \\ \cmidrule{4-7} 
                               &                            &                               & Inference  & 49,495 ms  & 11,494 ms  & 4.31\texttimes    \\
    \bottomrule
    \end{tabular}
    \label{tab:benchmarks}
\end{table}

\begin{table}[htb]
    \centering
    \caption{Training time and Accuracy (Acc.) for Covertype and Bank datasets, for the ARM CPU (1.2Ghz) and the FPGA (103MHz)}
    \begin{tabular}{crr|rrr}
    \toprule
              & \multicolumn{2}{c}{ARM CPU} & \multicolumn{3}{c}{FPGA}   \\ \cmidrule{2-6}
              & Acc.   & Time     & Acc. & Time       & Speedup    \\ \midrule
    Bank      & 88.3\%    & 202 ms   & 88.3\%  & 8,525 ms   & 0.02\texttimes    \\ \midrule
    Covertype & 72.2\%    & 9,712 ms & 63.7\%  & 374,600 ms & 0.03\texttimes    \\
    \bottomrule
    \end{tabular}
    \label{tab:uci_datasets}
\end{table}

\section{Related Work}

Kulaga et al. \cite{Kuaga2014FPGAHls} present an \ac{HLS} decision tree ensemble solution for inference tasks. The results achieved are competitive regarding performance when compared to the ARM core present in the tested SoC. 
However, the design is highly dependent on the number of trees and corresponding depths, as a change in ensemble parameters requires re-tuning multiple pragmas.
As we have also seen, an unavoidable sequential portion of the algorithm is the sample sorting through the tree structure.  
Unlike our approach, the number of trees in an ensemble is hardcoded into the synthesised kernel. In contrast, by having one or more synthesised training/inference methods (for different hyper-parameters), we can deploy \emph{N} instances of such circuits and process a runtime allocated number of trees.

As previously stated, the work on this paper builds on  Lin et al. \cite{Lin2019TowardsFPGA} work. However, their implementation is closed-source and done in Verilog, which excludes native execution on CPUs. Also, as the work was developed for a datacenter-class FPGA device, the implementation is very resource intensive and thus not suitable for small devices such as the ones used on embedded systems.

InAccel\footnote{\emph{InAccel, 2019, XGBoost Exact Updater IP core, } https://github.com/inaccel/xgboost} provides an HLS implementation of the XGBoost learning algorithm, which is also based on decision trees. For a dataset of 65k points, 5 classes, and 128 features, the training time is 2.7 seconds. This is significantly faster than our performance for similarly sized datasets, but InAccel's implementation targets server-grade FPGA accelerator boards (including multi-board setups), while we target the embedded domain. However, the potential for HLS FPGA acceleration of decision tree algorithms is demonstrated, given expert optimisation of the code for HLS.

\section{Conclusions}

We presented a flexible and scalable implementation of a Hoeffding Tree compatible with HLS tools%
\footnote{https://github.com/Sleepy105/Hoeffding-Tree/tree/fpt21}
We performed a functional validation of the tree design, against software execution, by implementation on chip on a Xilinx ZCU102. We provide a evaluation of the design's resource usage for multiple template parameter values (i.e., maximum tree size, number of sample attributes, number of clusters, and number of dataset samples), as well as execution time versus an ARM Cortex-A53 processor. The resource requirements of the tree do not scale significantly with problem size, although further HLS optimisations such as unrolling remain unexplored. Even so, we outperform the ARM by 8.3x times for largest dataset for the inference task, while being 8.6x slower during training. As future work, we envision the use of tree ensembles, and the partitioning of training and inference task between software and hardware based on problem size. 

\section*{Acknowledgments}

This work was supported by the PEPCC project (PTDC\slash EEI-HAC\slash 30848\slash 2017), financed by Fundação para a Ciência e Tecnologia (FCT).

\bibliographystyle{IEEEtran}
\bibliography{main}

\begin{thebibliography}{10}
\providecommand{\url}[1]{#1}
\csname url@samestyle\endcsname
\providecommand{\newblock}{\relax}
\providecommand{\bibinfo}[2]{#2}
\providecommand{\BIBentrySTDinterwordspacing}{\spaceskip=0pt\relax}
\providecommand{\BIBentryALTinterwordstretchfactor}{4}
\providecommand{\BIBentryALTinterwordspacing}{\spaceskip=\fontdimen2\font plus
\BIBentryALTinterwordstretchfactor\fontdimen3\font minus
  \fontdimen4\font\relax}
\providecommand{\BIBforeignlanguage}[2]{{%
\expandafter\ifx\csname l@#1\endcsname\relax
\typeout{** WARNING: IEEEtran.bst: No hyphenation pattern has been}%
\typeout{** loaded for the language `#1'. Using the pattern for}%
\typeout{** the default language instead.}%
\else
\language=\csname l@#1\endcsname
\fi
#2}}
\providecommand{\BIBdecl}{\relax}
\BIBdecl

\bibitem{XilinxInc.2020VivadoSynthesis}
\BIBentryALTinterwordspacing
{Xilinx Inc.}, ``{Vivado High-Level Synthesis},'' Online, Tech. Rep., 2020.
  [Online]. Available:
  \url{https://www.xilinx.com/products/design-tools/vivado/integration/esl-design.html}
\BIBentrySTDinterwordspacing

\bibitem{Quinlan1983LearningGames}
\BIBentryALTinterwordspacing
J.~R. Quinlan, ``{Learning Efficient Classification Procedures and Their
  Application to Chess End Games},'' \emph{Machine Learning}, pp. 463--482,
  1983. [Online]. Available:
  \url{https://link.springer.com/chapter/10.1007/978-3-662-12405-5\%5F15}
\BIBentrySTDinterwordspacing

\bibitem{UTGOFF1988ID5:ID3}
\BIBentryALTinterwordspacing
P.~E. Utgoff, ``{ID5: An Incremental ID3},'' in \emph{Machine Learning
  Proceedings 1988}.\hskip 1em plus 0.5em minus 0.4em\relax Elsevier, 1 1988,
  pp. 107--120. [Online]. Available:
  \url{https://linkinghub.elsevier.com/retrieve/pii/B9780934613644500177}
\BIBentrySTDinterwordspacing

\bibitem{Utgoff1989IMPROVEDLEARNING}
\BIBentryALTinterwordspacing
------, ``{Improved Training Via Incremental Learning},'' in \emph{Proceedings
  of the Sixth International Workshop on Machine Learning}.\hskip 1em plus
  0.5em minus 0.4em\relax Elsevier, 1 1989, pp. 362--365. [Online]. Available:
  \url{https://linkinghub.elsevier.com/retrieve/pii/B9781558600362500928}
\BIBentrySTDinterwordspacing

\bibitem{Utgoff1997DecisionRestructuring}
\BIBentryALTinterwordspacing
P.~E. Utgoff, N.~C. Berkman, and J.~A. Clouse, ``{Decision Tree Induction Based
  on Efficient Tree Restructuring},'' \emph{Machine Learning 1997 29:1},
  vol.~29, no.~1, pp. 5--44, 1997. [Online]. Available:
  \url{https://link.springer.com/article/10.1023/A:1007413323501}
\BIBentrySTDinterwordspacing

\bibitem{Domingos2000MiningStreams}
\BIBentryALTinterwordspacing
P.~Domingos and G.~Hulten, ``{Mining high-speed data streams},'' in
  \emph{Proceeding of the Sixth ACM SIGKDD International Conference on
  Knowledge Discovery and Data Mining}.\hskip 1em plus 0.5em minus 0.4em\relax
  New York, New York, USA: Association for Computing Machinery (ACM), 2000, pp.
  71--80. [Online]. Available:
  \url{http://portal.acm.org/citation.cfm?doid=347090.347107}
\BIBentrySTDinterwordspacing

\bibitem{Lin2019TowardsFPGA}
\BIBentryALTinterwordspacing
Z.~Lin, S.~Sinha, and W.~Zhang, ``{Towards Efficient and Scalable Acceleration
  of Online Decision Tree Learning on FPGA},'' in \emph{2019 IEEE 27th Annual
  International Symposium on Field-Programmable Custom Computing Machines
  (FCCM)}.\hskip 1em plus 0.5em minus 0.4em\relax IEEE, 4 2019, pp. 172--180.
  [Online]. Available: \url{https://ieeexplore.ieee.org/document/8735508}
\BIBentrySTDinterwordspacing

\bibitem{Pfahringer2008HandlingTrees}
\BIBentryALTinterwordspacing
B.~Pfahringer, G.~Holmes, and R.~Kirkby, ``{Handling Numeric Attributes in
  Hoeffding Trees},'' \emph{Lecture Notes in Computer Science (including
  subseries Lecture Notes in Artificial Intelligence and Lecture Notes in
  Bioinformatics)}, vol. 5012 LNAI, pp. 296--307, 2008. [Online]. Available:
  \url{https://link.springer.com/chapter/10.1007/978-3-540-68125-0\%5F27}
\BIBentrySTDinterwordspacing

\bibitem{Althoff2017AnTesting}
\BIBentryALTinterwordspacing
A.~Althoff and R.~Kastner, ``{An Architecture for Learning Stream Distributions
  with Application to RNG Testing},'' \emph{Proceedings - Design Automation
  Conference}, vol. Part 128280, 6 2017. [Online]. Available:
  \url{http://dx.doi.org/10.1145/3061639.3062199}
\BIBentrySTDinterwordspacing

\bibitem{Barbareschi2021AdvancingStudy}
\BIBentryALTinterwordspacing
M.~Barbareschi, S.~Barone, and N.~Mazzocca, ``{Advancing synthesis of decision
  tree-based multiple classifier systems: an approximate computing case
  study},'' \emph{Knowledge and Information Systems 2021 63:6}, vol.~63, no.~6,
  pp. 1577--1596, 4 2021. [Online]. Available:
  \url{https://link.springer.com/article/10.1007/s10115-021-01565-5}
\BIBentrySTDinterwordspacing

\bibitem{Kuaga2014FPGAHls}
\BIBentryALTinterwordspacing
R.~Ku{\l}aga and M.~Gorgo{\'{n}}, ``{FPGA Implementation of Decision Trees and
  Tree Ensembles for Character Recognition in Vivado Hls},'' \emph{Image
  Processing {\&} Communications}, vol.~19, no. 2-3, pp. 71--82, 9 2014.
  [Online]. Available:
  \url{https://www.sciendo.com/article/10.1515/ipc-2015-0012}
\BIBentrySTDinterwordspacing

\end{thebibliography}

\end{document}